\documentclass{article}

\usepackage{proceedings}
\usepackage{latexsym}
\usepackage{url}
\usepackage{amstext,amssymb}

\usepackage{times}
\usepackage{logic}

\newcommand{\INT}[1]{\mathit{INT}_{#1}}
\newcommand{\tr}[0]{\sigma}
\newcommand{\DLV}[0]{\texttt{DLV}}
\newcommand{\SM}[0]{\texttt{Smodels}}
\newcommand{\aux}[1]{\gamma(#1)}

\newcommand{\nnf}[1]{{\cal T}(#1)}
\renewcommand{\nnf}[1]{\nu(#1)}

\newcommand{\asal}[2]{\mathit{AS}_{#1}(#2)}

\newcommand{\prgm}{\mathcal{P}}
\renewcommand{\prgm}{\mathit{NLP}}

\newcommand{\dlp}{\mathcal{D}}
\renewcommand{\dlp}{\mathit{DLP}}
\newcommand{\gdlp}{\mathcal{D}^{g}}
\renewcommand{\gdlp}{\mathit{GDLP}}
\newcommand{\nnfprgm}{\prgm^{\mathit{nnf}}}
\newcommand{\gdlpht}{\gdlp^{\mathit{ht}}}

\newcommand{\LPif}{\leftarrow}
\newcommand{\lang}{\ensuremath{{\cal L}_{\cal A}}}

\newcommand{\al}{\ensuremath{{\cal A}}}

\newcommand{\F}{\ensuremath{{\cal F}}}
\renewcommand{\F}{\ensuremath{{\cal I}}}
\newcommand{\G}{\ensuremath{{\cal J}}}

\newcommand{\Iht}{\ensuremath{\langle I_H, I_T \rangle}}
\newcommand{\I}[0]{\ensuremath{\nu}}
\newcommand{\Ivo}[0]{\I}
\newcommand{\val}[2]{\Ivo_{#1}({#2})}
\newcommand{\valF}[2]{\val{\F}{#1,#2}}

\newcommand{\commadots}[0]{,\ldots ,}
\newcommand{\iec}[0]{i.e.,\ }

\newcommand{\egc}[0]{e.g.,\ }

\newcommand{\var}[1]{\ensuremath{\mathit{var(#1)}}}
\newtheorem{sublemma}{Sublemma}

\newcommand{\lab}{{L}}
\renewcommand{\lab}{\mathbf{L}}

\newcommand{\labf}[1]{{\lab}_{#1}}

\renewcommand{\var}[1]{\ensuremath{\mathit{var}(#1)}}

\newtheorem{definition}{Definition}
\newtheorem{proposition}{Proposition}
\newtheorem{theorem}{Theorem}
\newtheorem{lemma}{Lemma}
\newtheorem{corollary}{Corollary}

\author{%
  { \bf David Pearce}\\
European Commission,\\
DG Information Society -- F1\\
BU33 3/58, Rue de le Loi 200,\\
B-1049 Brussels \\
\texttt{David.Pearce@cec.eu.int}
\And
{\bf Vladimir Sarsakov},
{\bf Torsten Schaub\thanks{\ \ Affiliated with the School of
Computing Science at
Simon Fraser University,
Burnaby, Canada.}}\\
Institut f\"ur Informatik,
Universit\"at Potsdam,\\
Postfach 90 03 27,
D--14439 Potsdam,
 Germany\\
\texttt{sarsakov@rz.uni-potsdam.de},\\
\texttt{torsten@cs.uni-potsdam.de}\\
\AND
{\bf Hans Tompits},
{\bf Stefan Woltran}\\
Institut f\"ur Informationssysteme
184/3,\\ 
Technische Universit\"at Wien, \\
Favoritenstra{\ss}e~9--11,
A--1040 Vienna, Austria\\
\texttt{[tompits,stefan]@kr.tuwien.ac.at}
}

\title{\bf A Polynomial Translation
of Logic Programs with Nested Expressions into Disjunctive Logic Programs: %\\ 
Preliminary Report\thanks{\ \ This work was partially supported by
the Austrian Science Fund (FWF) under grants P15068-INF and  N Z29-INF.
The authors would like to thank Agata Ciabattoni for pointing out some relevant references.}}

\begin{document}

\maketitle

\begin{abstract}

Nested logic programs have recently been introduced in order to
allow for
arbitrarily nested formulas in the heads and the bodies of logic
program
rules under the answer sets semantics.
Nested expressions can be formed using conjunction, disjunction,
as well as
the negation as failure operator in an unrestricted
fashion.
This provides a very flexible and compact framework for knowledge
representation and reasoning.
Previous results show that nested logic programs can be transformed into
standard (unnested) disjunctive logic programs in an elementary way, applying 
the
negation as failure
operator to body literals only.
This is of great practical relevance since it allows us to
evaluate nested
logic programs by means of off-the-shelf disjunctive logic
programming
systems, like \DLV.
However, it turns out that this straightforward transformation
results in an
exponential blow-up in the worst-case, despite the fact that
complexity
results indicate that there is a polynomial translation among
both formalisms.
In this paper, we take up this challenge  and provide a polynomial
translation
of logic programs with nested expressions into disjunctive logic
programs.
Moreover, we show that this translation is modular and (strongly)
faithful.
We have implemented both the straightforward as well as our
advanced
transformation;
the resulting compiler serves as a front-end to \DLV\ and is publicly
available on
the Web.
\end{abstract}

\section{Introduction}

Lifschitz, Tang, and Turner~\cite{Lifschitz99} recently extended the answer
set semantics~\cite{Gelfond91} to a class of logic programs in which arbitrarily nested
formulas, formed from literals using negation as failure, conjunction, and
disjunction, constitute the heads and bodies of rules.
These so-called \emph{nested logic programs} generalise the well-known
classes of \emph{normal}, \emph{generalised}, \emph{extended}, and
\emph{disjunctive logic programs}, respectively.
Despite their syntactically much more restricted format, the latter classes are
well recognised as important tools for knowledge representation and reasoning.
This is reflected by the fact that several practicably relevant applications
have been developed recently using these types of programs (cf., \egc ~\cite{Lifschitz99a,BarUyan:2001,Gelfond:2001:DPS,HelNiem:2001}),
which in turn is largely fostered by the availability of efficient solvers for
the answer set semantics, most notably \DLV~\cite{Eiter97,Eiter98a} and
\SM~\cite{Niemelae97}.

In this paper, we are interested in utilising these highly performant solvers
for interpreting nested logic programs.
We address this problem by providing a translation of nested logic programs
into disjunctive logic programs.
In contrast to previous work, our translation is guaranteed to be polynomial
in time and space, as suggested by related complexity results~\cite{Pearce01}.
More specifically, we provide a translation, $\tr$, from nested logic programs
into disjunctive logic programs possessing the following properties:
\begin{itemize}
\item $\tr$ maps nested logic programs over an alphabet $\al_1$ into
  disjunctive logic programs over an alphabet $\al_2$, where
  $\al_1\subseteq\al_2$;
  
\item the size of $\tr(\Pi)$ is polynomial in the size of $\Pi$;

\item $\tr$ is \emph{faithful}, \iec for each program $\Pi$ over alphabet
  $\al_1$, there is a one-to-one correspondence between the answer sets of
  $\Pi$ and sets of form $I\cap\al_1$, where $I$ is an answer set of
  $\tr(\Pi)$; and

\item $\tr$ is \emph{modular}, \iec $\tr(\Pi\cup\Pi')=\tr(\Pi)\cup\tr(\Pi')$,
  for each program $\Pi,\Pi'$.

\end{itemize}
Moreover, we have implemented translation $\tr$, serving as a front-end for the
logic programming system \DLV. 

The construction of $\tr$ relies on the introduction of new \emph{labels},
abbreviating subformula occurrences.
This technique is derived from \emph{structure-preserving normal form
translations}~\cite{Tseitin70,Greenbaum86}, frequently employed in the context
of automated deduction (cf.~\cite{Baaz01} for an overview).
We use here a method adapted from a structure-preserving translation for
intuitionistic logic as described in~\cite{Mints94}.

Regarding the faithfulness of $\tr$, we actually provide a somewhat stronger
condition, referred to as \emph{strong faithfulness}, expressing that, for
any programs $\Pi$ and $\Pi'$ over alphabet $\al_1$, there is a one-to-one
correspondence between the answer sets of $\Pi\cup\Pi'$ and sets of form
$I\cap\al_1$, where $I$ is an answer set of $\tr(\Pi)\cup\Pi'$.
This condition means that we can add to a given program $\Pi$ \emph{any nested program} $\Pi'$ and still recover the answer sets of the combined program $\Pi\cup\Pi'$ from $\tr(\Pi)\cup\Pi'$;
in particular, 
for any
nested logic program $\Pi$, we may choose
to translate, in a semantics-preserving way, only an arbitrary \emph{program part}
 $\Pi_0\subseteq \Pi$ and
leave the remaining part $\Pi\setminus \Pi_0$ unchanged.
For instance, if $\Pi_0$ is already a disjunctive logic program, we do not
need to translate it again into another (equivalent) disjunctive logic
program.
Strong faithfulness is closely related to the concept of \emph{strong
  equivalence}~\cite{Lifschitz01} (see below).
 
In order to have a sufficiently general setting for our purposes, we base our
investigation on \emph{equilibrium logic}~\cite{Pearce97}, a generalisation 
of the answer set semantics for nested logic programs. 
Equilibrium logic is a form of minimal-model reasoning in the 
\emph{logic of here-and-there}, 
which is intermediate between classical logic and intuitionistic logic (the
logic of here-and-there is also known as \emph{G\"odel's three-valued logic}
in view of~\cite{Goedel32}).
As shown in \cite{Pearce97,Pearce99,Lifschitz01}, 
logic programs can be viewed as a special class of formulas in the logic of
here-and-there such that, for each program $\Pi$, the answer sets of $\Pi$ are
given by the equilibrium models of $\Pi$, where the latter $\Pi$ is viewed as
a set of formulas in the logic of here-and-there.

The problem of implementing nested logic programs has already been addressed in~\cite{Pearce01}, where (linear-time constructible) encodings of the basic
reasoning tasks associated with this language into quantified Boolean 
formulas are described.
These encodings provide a straightforward implementation for nested logic
programs by appeal to off-the-shelf solvers for quantified Boolean formulas
(like, \egc the systems proposed
in~\cite{Cadoli98,Feldmann00,Giunchiglia01a,Buening95,Letz01,Rintanen99a}).
Besides the encodings into quantified Boolean formulas, a further result of
\cite{Pearce01} is that nested logic programs possess the same worst-case
complexity as disjunctive logic programs, \iec the main reasoning tasks
associated with nested logic programs lie at the second level of the
polynomial hierarchy.
From this result it follows that nested logic programs can in turn be
efficiently reduced to disjunctive logic programs.
Hence, given such a reduction, solvers for the latter kinds of programs, 
like, \egc \DLV\ or \SM, can be
used to compute the answer sets of nested logic programs.
The main goal of this paper is to construct a reduction of this type.

Although results by Lifschitz, Tang, and Turner~\cite{Lifschitz99} (together with transformation rules given in~\cite{Janhunen01}) provide a method to translate nested logic programs into disjunctive ones, that approach suffers from the drawback of an
exponential blow-up of the resulting disjunctive logic programs in the worst case.
This is due to the fact that the ``{language-preserving}'' nature of that translation 
relies on distributivity laws yielding an exponential increase of program size whenever the given program contains rules whose heads
are in disjunctive normal form or whose bodies are in conjunctive normal form, 
and the respective expressions are not simple disjunctions or conjunctions of
literals.
Our translation, on the other hand, is always polynomial in the size of its input program.

Finally, we mention that structure-preserving normal form translations in the logic of
here-and-there are also studied, yet in much more general settings, by Baaz and
Ferm\"uller~\cite{Baaz95} as well as by H\"ahnle~\cite{Haehnle94}; there, whole
classes of finite-valued G\"odel logics are investigated.
Unfortunately, these normal form translations are not suitable for our purposes,
because they do not enjoy the particular form of programs required here.

\section{Preliminaries}

We deal with propositional languages and use the logical symbols
$\top$, $\bot$, $\neg$, $\vee$, $\wedge$, and 
$\IMPL$ 
to construct formulas in the standard way.
We write $\lang$ to denote a language over an alphabet
$\al$
of {\em propositional variables} or 
{\em atoms}.
Formulas are denoted by Greek lower-case letters (possibly with subscripts).
As usual, \emph{literals} are formulas of form $v$ or $\neg v$, where $v$ is some variable or one of $\top,\bot$.

Besides the semantical concepts introduced below, we also make use of the 
semantics of classical propositional logic.
By a \emph{$($classical\/$)$ interpretation}, $I$, we understand a set of variables.
Informally, a variable $v$ is true under $I$ iff $v\in I$.
The truth value of a formula $\phi$ under interpretation $I$, 
in the 
sense of classical propositional logic, 
is determined in the usual way.

%%%%%%%%%%%%%%%%%%%%%%%%%%%%%%%%%%%%%%%%%%%%%%%%%%%%%%%%%%%%%%%%%%%%%%%%%%%%
\subsection{Logic Programs}\label{sec:lp}
%%%%%%%%%%%%%%%%%%%%%%%%%%%%%%%%%%%%%%%%%%%%%%%%%%%%%%%%%%%%%%%%%%%%%%%%%%%%

The central objects of our investigation are 
logic programs with nested expressions, introduced by 
Lifschitz \emph{et al}.~\cite{Lifschitz99}.
 These kinds of programs
generalise normal logic programs
by allowing bodies and heads of rules to contain arbitrary Boolean
formulas.
For reasons of simplicity, we  
deal here  only with languages containing one kind of negation, however, 
corresponding to default negation.
The extension to the general case where strong negation is also permitted 
is straightforward and proceeds in the usual way.

We start with some basic notation. A formula  whose sentential connectives 
comprise only $\AND$, $\OR$, or 
$\neg$ is called an \emph{expression}. A rule, $r$, is an ordered pair of form
\[
H(r) \LPif B(r),
\]
where $B(r)$ and $H(r)$ are expressions.
$B(r)$ is called the \emph{body} of $r$ and $H(r)$ is the 
\emph{head} of $r$. 
We say that $r$ is a \emph{generalised disjunctive rule} if 
$B(r)$ is a conjunction of literals and $H(r)$ is 
a disjunction of literals;
$r$ is a \emph{disjunctive rule} iff it is a generalised 
disjunctive rule containing no negated atom in its head; 
finally, if $r$ is a rule containing no negation at all, then $r$ is 
called \emph{basic}.
A \emph{nested logic program}, or simply a \emph{program}, $\Pi$, is a 
finite set of rules.
$\Pi$ is a \emph{generalised disjunctive logic program} iff it contains only
generalised disjunctive rules.
Likewise, $\Pi$ is a \emph{disjunctive logic program} iff 
$\Pi$ contains only disjunctive rules, and $\Pi$ is 
\emph{basic} iff each rule in $\Pi$ is basic.
We say that $\Pi$ is a program \emph{over} alphabet $\al$ 
iff all atoms occurring in $\Pi$ are from $\al$.
The set of all atoms occurring in program $\Pi$ is denoted by $\var{\Pi}$. 
We use $\prgm_\al$ to denote the class of all nested logic programs over 
alphabet $\al$; furthermore, $\dlp_\al$ stands for the subclass 
of $\prgm_\al$ containing all disjunctive logic programs over~$\al$; 
and $\gdlp_\al$ is the class of all generalised disjunctive 
logic programs over~$\al$.
Further classes of programs are introduced in Section~\ref{sec:poly}.

In what follows, we associate to each rule $r$ a corresponding 
formula $\hat{r}=B(r)\IMPL H(r)$ and, accordingly, to each 
program $\Pi$ a corresponding set of formulas 
$\hat{\Pi}=\{\hat{r}\mid r\in\Pi\}$.

Let  $\Pi$ be a basic program over $\al$ and $I\subseteq\al$ 
a (classical) interpretation.
We say that $I$ is a \emph{model} of 
$\Pi$ iff it is a model of 
the associated set $\hat{\Pi}$ of formulas.
Furthermore, given an (arbitrary) program $\Pi$ over $\al$, 
the \emph{reduct}, $\Pi^I$, of $\Pi$ 
with respect to $I$ is the basic program obtained from $\Pi$ by 
replacing every occurrence of an expression $\neg \psi$ in 
$\Pi$ which is not in the scope of any other negation by $\bot$ 
if $\psi$ is true under $I$, and by $\top$ otherwise. 
$I$ is an \emph{answer set} (or \emph{stable model}) of $\Pi$ iff it is a minimal 
model (with respect to set inclusion) of the reduct $\Pi^I$.
The collection of all answer sets of $\Pi$ is denoted by $\asal{\al}{\Pi}$.

Two logic programs, $\Pi_1$ and $\Pi_2$, are \emph{equivalent} 
iff they possess the same answer sets. 
Following 
Lifschitz \emph{et al}.~\cite{Lifschitz01}, we call $\Pi_1$ 
and $\Pi_2$ \emph{strongly equivalent} iff,
for every program $\Pi$, $\Pi_1\cup\Pi$ and $\Pi_2\cup\Pi$ are equivalent.

%%%%%%%%%%%%%%%%%%%%%%%%%%%%%%%%%%%%%%%%%%%%%%%%%%%%%%%%%%%%%%%%%%%%%%%%%%
\subsection{
Equilibrium Logic
}
%%%%%%%%%%%%%%%%%%%%%%%%%%%%%%%%%%%%%%%%%%%%%%%%%%%%%%%%%%%%%%%%%%%%%%%%%%

Equilibrium logic is an approach to 
nonmonotonic reasoning that generalises the answer set 
semantics for logic programs. 
We use this particular formalism because it offers a convenient 
logical language for dealing with  logic programs under 
the answer set semantics.
It is defined in terms of the logic of here-and-there, 
which is intermediate between classical logic and intuitionistic logic.
Equilibrium logic was introduced in~\cite{Pearce97} and 
further investigated in \cite{Pearce99}; proof theoretic 
studies of the logic can be found in~\cite{Pearce00,Pearce00a}. 

Generally speaking, the logic of here-and-there is an important tool 
for analysing various properties of logic programs. 
For instance, as shown in \cite{Lifschitz01}, the problem of 
checking whether two logic programs are strongly equivalent 
can be expressed in terms of the logic of here-and-there (cf.\ 
Proposition~\ref{prop:strong} below).

The semantics of the logic of here-and-there is defined
by means of two worlds, $H$ and $T$, called ``here'' and ``there''.
It is assumed that there is a total order, $\leq$, defined between 
these worlds such that $\leq$ is reflexive and $H \leq T$. 
As in ordinary Kripke semantics for intuitionistic logic, we can 
imagine that in each world a set of atoms is verified and that, 
once verified ``here'', an atom remains verified ``there''. 

Formally, by an \emph{HT-interpretation}, $\F$, we understand an 
ordered pair $\Iht$ of sets of atoms such that $I_H \subseteq I_T$.
We say that $\F$ is an HT-interpretation \emph{over} $\al$ if $I_T\subseteq\al$. 
The set of all HT-interpretations over $\al$ is denoted by $\INT{\al}$.
An HT-interpretation $\Iht$ is \emph{total} if $I_H = I_T$.

The truth value, $\valF{w}{\phi}$, of a formula $\phi$ at a 
world $w\in\{H,T\}$ in an HT-interpretation $\F=\Iht$ is recursively 
defined as follows:
\begin{enumerate}
\item if $\phi=\top$, then $\valF{w}{\phi} = 1$;

\item if $\phi=\bot$, then $\valF{w}{\phi} = 0$;

\item if $\phi=v$ is an atom, then $\valF{w}{\phi} = 1$ if
      $v\in I_w$, otherwise $\valF{w}{\phi} = 0$;

\item if $\phi=\neg \psi$, then
      $\valF{w}{\phi} = 1$ if,  for every world $u$ with $w\leq u$,
      $\valF{u}{\psi} = 0$,
otherwise $\valF{w}{\phi} = 0$;
\item if $\phi=(\phi_1\AND \phi_2)$, then
      $\valF{w}{\phi}=1$ if $\valF{w}{\phi_1}=1$ and $\valF{w}{\phi_2}=1$,
otherwise $\valF{w}{\phi}=0$;

\item if $\phi=(\phi_1\OR \phi_2)$, then
      $\valF{w}{\phi}=1$ if $\valF{w}{\phi_1}=1$ or $\valF{w}{\phi_2}=1$,
otherwise $\valF{w}{\phi}=0$;

\item if $\phi=(\phi_1\IMPL \phi_2)$, then
      $\valF{w}{\phi} = 1$ if for every world $u$ with $w\leq u$,
        $\valF{u}{\phi_1}=0$ or $\valF{u}{\phi_2}=1$,
otherwise $\valF{w}{\phi} = 0$.
\end{enumerate}

We say that $\phi$ is \emph{true under $\F$ in $w$} iff $\valF{w}{\phi}=1$,
otherwise $\phi$ is \emph{false under $\F$ in $w$}.
An HT-interpretation $\F=\Iht$ \emph{satisfies} $\phi$, or $\F$ is an 
\emph{HT-model} 
of $\phi$, iff $\valF{H}{\phi}=1$.
If $\phi$ is true under any HT-interpretation, then $\phi$ is 
\emph{valid in the logic of here-and-there}, or simply \emph{HT-valid}.

Let $S$ be a set of formulas. An HT-interpretation $\F$ is an HT-model 
of $S$ iff $\F$ is an HT-model of each element of $S$. 
We say that $\F$ is an HT-model of a \emph{program}
$\Pi$ iff $\F$ is an HT-model of $\hat{\Pi}=
\{B(r)\IMPL H(r)\mid r\in\Pi\}$.

Two sets of formulas are \emph{equivalent in the logic 
of here-and-there}, or \emph{HT-equivalent}, iff they 
possess the same HT-models.
Two formulas, $\phi$ and $\psi$, are HT-equivalent iff 
the sets $\{\phi\}$ and $\{\psi\}$ are HT-equivalent.

It is easily seen that any HT-valid formula is valid in classical logic, 
but the converse does not always hold. For instance, $p\OR\neg p$ and 
$\neg\neg p\IMPL p$ are valid in classical logic but not in the logic 
of here-and-there as the pair $\langle \emptyset,\{p\}\rangle$ is not 
an HT-model for either of these formulas.

\emph{Equilibrium logic} can be seen as a particular type of reasoning 
with minimal HT-models. Formally, an \emph{equilibrium model} of a 
formula $\phi$ is a total HT-interpretation $\langle I,I \rangle$ 
such that (i) $\langle I,I \rangle$ is an HT-model of $\phi$, 
and (ii) for every proper subset $J$ of $I$, $\langle J,I \rangle$ 
is not an HT-model of $\phi$.

The following result establishes the close connection 
between equilibrium models and answer sets, showing 
that answer sets are actually a special case of equilibrium models:

 \begin{proposition}[\cite{Pearce97,Lifschitz01}] \label{prop:eqmsm}
For any program $\Pi$, $I$ is an answer set of\/ $\Pi$
iff
$\langle I, I \rangle$ is an equilibrium model of~$\hat{\Pi}$.
\end{proposition}

Moreover, HT-equivalence was shown to capture the notion of strong equivalence between logic programs:

\begin{proposition}[\cite{Lifschitz01}] \label{prop:strong}
Let $\Pi_1$ and $\Pi_2$ be programs, and let 
$\hat{\Pi}_i=\{B(r)\IMPL H(r)\mid r\in\Pi_i\}$, for $i=1,2$. 
Then, $\Pi_1$ and $\Pi_2$ are strongly equivalent iff\/
$\hat{\Pi}_1$ and $\hat{\Pi}_2$ are equivalent in the 
logic of here-and-there.
\end{proposition} 

Recently, de Jongh and Hendriks~\cite{jongh-hendriks-01} 
have extended Proposition~\ref{prop:strong}
by showing that for nested programs strong equivalence is
characterised
precisely by equivalence in all intermediate logics lying between
here-and-there (upper bound) and the logic KC of weak excluded
middle (lower
bound) which is axiomatised by intuitionistic logic together with
the schema
$\neg \varphi \vee \neg \neg \varphi$.

We require the following additional concepts.
By an \emph{HT-literal}, $l$, we understand a formula of form $v$, 
$\neg v$, or $\neg \neg v$, 
where $v$ is a propositional atom or one of $\top$, $\bot$.
Furthermore, a formula is in 
\emph{here-and-there negational normal form}, or \emph{HT-NNF}, 
if it is made up of HT-literals, conjunctions and disjunctions.
Likewise, we say that a \emph{program} is in HT-NNF iff all heads 
and bodies of rules in the program are in HT-NNF.

Following~\cite{Lifschitz99}, every expression $\phi$ can 
effectively be transformed into an expression $\psi$ in 
HT-NNF possessing the same HT-models as $\phi$.
In fact, we have the following property: 

\begin{proposition}\label{prop:NNF}
Every expression $\phi$ is HT-equivalent to
an expression $\nnf{\phi}$ in HT-NNF, where  $\nnf{\phi}$ is 
constructible in polynomial time from $\phi$, satisfying the 
following conditions, for each expression $\varphi,\psi$:
\begin{enumerate}
\item $\nnf{\varphi}=\varphi$, if $\varphi$ is an HT-literal;
\item $\nnf{\neg\neg\neg\varphi}=\nnf{\neg\varphi}$; 
\item $\nnf{\varphi\circ\psi}=\nnf{\varphi}\circ\nnf{\psi}$, for $\circ\in\{\wedge,\vee\}$; 
\item $\nnf{\neg(\varphi\AND\psi)}=\nnf{\neg\varphi}\OR\nnf{\neg\psi}$;
\item $\nnf{\neg(\varphi\OR\psi)}=\nnf{\neg\varphi}\AND\nnf{\neg\psi}$.
\end{enumerate}
\end{proposition}

\section{Faithful Translations
}\label{sec:neq}

Next, we introduce the general requirements we impose 
on our desired translation from nested logic programs 
into disjunctive logic programs.
The following definition is central:

%%%%%%%%%%%%%%%%%%%%%%%%%%%%%%%%%%%%%%%%%%%%%definition
\begin{definition}\label{def:eq}
Let $\al_1$ and $\al_2$ be two alphabets such that 
$\al_1\subseteq\al_2$, and, for $i=1,2$, let 
$S_i\subseteq\prgm_{\al_i}$ be a class of nested 
logic programs closed under unions.\footnote{A class 
$S$ of sets is closed under unions providing $A,B\in S$ 
implies $A\cup B\in S$.}
Then, a function $\rho:S_1\rightarrow S_2$ is
\begin{enumerate}
\item\label{def:faithful} \emph{polynomial} iff, for all 
programs $\Pi\in S_1$, the time required to 
compute $\rho(\Pi)$ is polynomial in the size of $\Pi$;

\item \emph{faithful} iff, for all programs  
$\Pi\in S_1$, 
\[
\asal{\al_1}{\Pi}=\{I\cap{\al_1}\mid I\in\asal{\al_2}{\rho(\Pi)}\};
\]

\item \emph{strongly faithful} iff, for all programs  
$\Pi\in S_1$ and all programs $\Pi'\in \prgm_{\al_1}$, 
\[
\asal{\al_1}{\Pi\cup\Pi'}=\{I\cap\al_1\mid I 
\in\asal{\al_2}{\rho(\Pi)\cup\Pi'}\};
\]
and
\item \emph{modular} iff, for all programs $\Pi_1,
\Pi_2\in S_1$, 
\[
\rho(\Pi_1\cup\Pi_2)=\rho(\Pi_1)\cup\rho(\Pi_2).
\]

\end{enumerate}
\end{definition}

In view of the requirement that $\al_1\subseteq\al_2$, the 
general functions considered here may introduce new atoms.
Clearly, if the given function is polynomial, the number 
of newly introduced atoms is also polynomial.
Faithfulness guarantees that we can recover the stable 
models of the input program from the translated program.
Strong faithfulness, on the other hand, states that we 
can add to a given program $\Pi$  \emph{any nested logic program} 
$\Pi'$  and still retain, up to the original language, the 
semantics of the combined program $\Pi\cup\Pi'$ from $\rho(\Pi)\cup\Pi'$.
Finally, modularity enforces that we can translate programs rule by rule.

It is quite obvious that any strongly faithful function is also faithful.
Furthermore, strong faithfulness of function $\rho$ implies that, 
for a given program $\Pi$,  we can translate \emph{any program part} 
$\Pi_0$ of $\Pi$ whilst leaving the remaining part 
$\Pi\setminus\Pi_0$ unchanged, and determine the semantics 
of $\Pi$ from $\rho(\Pi_0)\cup (\Pi\setminus\Pi_0)$.
As well, for any function of form $\rho:\prgm_\al\rightarrow\prgm_\al$, 
strong faithfulness of $\rho$ is equivalent to
the condition that $\Pi$ and $\rho(\Pi)$ are strongly 
equivalent, for any $\Pi\in\prgm_\al$.
Hence, strong faithfulness generalises strong equivalence.

Following~\cite{Janhunen98,Janhunen01}, 
we say that a 
function $\rho$ as in Definition~\ref{def:eq}
is \emph{PFM}, or that $\rho$ is a \emph{PFM-function}, 
iff it is polynomial, faithful, and modular.
Analogously, we call $\rho$ \emph{PSM}, or a \emph{PSM-function}, 
iff it is polynomial, strongly faithful, and modular.

It is easy to see that the composition of two PFM-functions is 
again a PFM-function; and likewise for PSM-functions.
Furthermore, since any PSM-function is also PFM, in the following 
we focus on PSM-functions.
In fact, in the next section, we construct a function 
$\tr:\prgm_{\al_1}\rightarrow \dlp_{\al_2}$ (where $\al_2$ 
is a suitable extension of $\al_1$) which is PSM.

Next, we discuss some sufficient conditions guaranteeing that 
certain classes of functions are strongly faithful.
We start with the following concept.

\begin{definition}\label{def:mono}
Let $\rho:\prgm_{\al_1}\rightarrow \prgm_{\al_2}$ be a 
function such that $\al_1\subseteq\al_2$, and let $\INT{\al_i}$ 
be the class of all HT-interpretations over $\al_i$ $(i=1,2)$.
 
Then, the function 
$\alpha_\rho:\INT{\al_1}\times \prgm_{\al_1}\rightarrow \INT{\al_2}$
is called a \emph{$\rho$-associated HT-embedding} iff, for each 
HT-interpretation $\F=\Iht$ over $\al_1$, each $\Pi\in \prgm_{\al_1}$, 
and each 
$w\in\{H,T\}$, $J_w\cap\al_1=I_w$ and $J_w\setminus\al_1\subseteq\var{\rho(\Pi)}$, 
where $\alpha_\rho(\F,\Pi)=\langle J_H,J_T\rangle$.

Furthermore, for any $G\subseteq\INT{\al_1}$ and any $\Pi\in \prgm_{\al_1}$,
we define $\alpha_\rho(G,\Pi)=\{\alpha_\rho(\F,\Pi)\mid \F\in G\}$.
\end{definition}

Intuitively, a $\rho$-associated HT-embedding transforms 
HT-interpretations over the input alphabet $\al_1$ of $\rho$ 
into HT-interpretations over the output alphabet $\al_2$ of 
$\rho$ such that the truth values of the atoms in $\al_1$ are retained.
The following definition strengthens these kinds of mappings:

%%%%%%%%%%%%%%%%%%%%%%%%%%%%%%%%%%%%%%%%%%%%%%%%%%definition
\begin{definition}\label{def:htnew}
Let $\rho$ be as in Definition~\ref{def:mono}, and let 
$\alpha_\rho$ be a $\rho$-associated HT-embedding. 
We say that $\alpha_\rho$ is a \emph{$\rho$-associated 
HT-homomorphism} if, for any $\F,\F'\in\INT{\al_1}$ and any 
$\Pi\in \prgm_{\al_1}$, the following conditions hold:
\begin{enumerate}
\item
$\F$ is an HT-model of\/ $\Pi$ iff 
$\alpha_\rho(\F,\Pi)$ is an HT-model of $\rho(\Pi)$;

 \item
$\F$ is total iff $\alpha_\rho(\F,\Pi)$ is total; 

\item
if $\F=\Iht$ and $\F'=\langle I'_H,I'_T\rangle$ are 
HT-models of\/ 
$\Pi$, then $I_H\subset I'_H$ and $I_T=I'_T$ holds 
precisely if $J_H\subset J_H'$
 and $J_T=J_T'$, for 
$\alpha_\rho(\F,\Pi)=\langle J_H,J_T\rangle$ 
and $\alpha_\rho(\F',\Pi)=\langle J_H',J_T'\rangle$;
and

\item
an HT-interpretation $\G$ over $\var{\rho(\Pi)}$ 
is an HT-model of $\rho(\Pi)$ only if 
$\G\in\alpha_\rho(\INT{\al_1},\Pi)$.
\end{enumerate}

\end{definition}

Roughly speaking, $\rho$-associated HT-homomorphisms retain the 
relevant  properties of HT-inter\-pretations for being 
equilibrium models with respect to transformation $\rho$.
More specifically, the first three conditions take semantical 
and set-theoretical properties into account, respectively, whilst 
the last one expresses a specific ``closure condition''.
The inclusion of the latter requirement is explained by observation that
the first
three conditions alone are not sufficient to 
exclude the possibility that there may exist some equilibrium 
model $\F$ of $\Pi$ such that $\alpha_\rho(\F,\Pi)$ is not 
an equilibrium model of  $\rho(\Pi)$.
The reason for this is that 
the set $\alpha_\rho(\INT{\al_1},\Pi)$, comprising the 
images of all HT-interpretations over $\al_1$ under 
$\alpha_\rho$ with respect to program $\Pi$, does, 
in general, not cover \emph{all} HT-interpretations over 
$\var{\rho(\Pi)}$.
Hence, for a general $\rho$-associated HT-embedding $\alpha_\rho(\cdot,\cdot)$,
there may exist some HT-model of $\rho(\Pi)$ 
which is not included in $\alpha_\rho(\INT{\al_1},\Pi)$ 
preventing $\alpha_\rho(\F,\Pi)$ from being an equilibrium 
model of $\rho(\Pi)$ albeit $\F$ is an equilibrium model of $\Pi$.
The addition of the last condition in Definition~\ref{def:htnew}, 
however, excludes this possibility, ensuring  that
all relevant HT-interpretations required for checking whether 
$\alpha_\rho(\F,\Pi)$ is an equilibrium model of $\rho(\Pi)$ 
are indeed considered.
The following result can be shown:

%%%%%%%%%%%%%%%%%%%%%%%%%%%%%%%%%%%%%%%%%%%theorem
\begin{lemma}\label{lemma:faithfulness}
For any 
function 
$\rho:\prgm_{\al_1}\rightarrow \prgm_{\al_2}$ 
with 
$\al_1\subseteq\al_2$, 
if there is some $\rho$-associated HT-homo\-morphism, 
then $\rho$ is faithful.
\end{lemma}

From this, we obtain the following property:

%%%%%%%%%%%%%%%%%%%%%%%%%%%%%%%%%%%%%%%%%%%theorem
\begin{theorem}\label{thm:strong-faithfulness}
Under the circumstances of Lemma~\ref{lemma:faithfulness}, 
if $\rho$ is modular and there is some $\rho$-associated 
HT-homomor\-phism, then $\rho$ is strongly faithful.
\end{theorem}

We make use of the last result for showing that the 
translation from nested logic programs into disjunctive 
logic programs, as discussed next, is PSM.

\section{Main Construction}
\label{sec:poly}

In this section, we show how logic programs with nested expressions 
can be efficiently mapped to disjunctive logic programs, preserving 
the semantics of the respective programs.
Although results by Lifschitz~\emph{et al.}~\cite{Lifschitz99} already provide
a reduction of nested logic programs into disjunctive ones (by employing additional transformation steps as given in~\cite{Janhunen01}), that 
method is exponential in the worst case.
This is due to the fact that the transformation relies on 
distributive laws, yielding an exponential increase of program 
size whenever the given program contains  rules whose heads are 
in disjunctive normal form or whose bodies are in conjunctive normal form, and the respective expressions are not simple disjunctions or conjunctions of HT-literals.

To avoid such an exponential blow-up, our technique is based on 
the introduction of new atoms, called \emph{labels}, abbreviating 
subformula occurrences.
This method is derived from structure-preserving normal form 
translations~\cite{Tseitin70,Greenbaum86}, which are frequently 
applied in the context of automated reasoning (cf., \egc 
\cite{Baaz95,Haehnle94} for general investigations about 
structure-preserving normal form translation in finite-valued 
G\"odel logics, and \cite{Egly96,Egly97} 
for proof-theoretical issues of such translations for classical 
and intuitionistic logic).
In contrast to theorem proving applications, where the main 
focus is to provide translations which are satisfiability 
(or, alternatively, validity) equivalent, here we are 
interested in somewhat stronger equivalence properties, 
viz.\ in the \emph{reconstruction of the answer sets} 
of the original programs from the translated ones, which 
involves also an adequate handling of additional minimality criteria. 

The overall structure of our translation can be described as follows.
Given a nested logic program $\Pi$, we perform the following steps:

\begin{enumerate}
\item\label{step:1} For each $r\in\Pi$, transform $H(r)$ and 
$B(r)$ into HT-NNF;

\item\label{step:2} translate the program into a program containing 
only rules with conjunctions of HT-literals in their bodies and 
disjunctions of HT-literals in their heads;

\item\label{step:4} eliminate double negations in bodies and heads; and

\item\label{step:5} transform the resulting program into a disjunctive 
logic program, \iec make all heads negation free.  
\end{enumerate}

Steps~\ref{step:1} 
and \ref{step:4} are realised by using properties of logic programs 
as described in~\cite{Lifschitz99}; 
Step~\ref{step:2} represents the central part of our construction;  
and Step~\ref{step:5} exploits a procedure due to Janhunen~\cite{Janhunen01}.

\bigskip
In what follows, for any alphabet $\al$, we define the following new and disjoint alphabets:
\begin{itemize}

\item a set $\al_\lab
= \{ \labf{\phi} \mid \phi\in\lang\}$ of labels; and

\item a set $\bar{\al}=\{\overline{p}\mid p\in\al\}$
of atoms representing negated atoms.

\end{itemize}

Furthermore, $\nnfprgm_{\al}$ is the class of all nested 
logic programs over $\al$ which are in  HT-NNF, and $\gdlpht_{\al}$ 
is the class of all programs over $\al$  which are defined like generalised 
logic programs, except that HT-literals may occur in rules 
instead of ordinary literals.

We assume that for each of the above construction stages, Step~$i$ 
is realized by a corresponding function $\tr_i(\cdot)$ ($i=1\commadots 4$).
The overall transformation is then described by the composed function 
$\tr=\tr_4\circ\tr_3\circ\tr_2\circ\tr_1$, which is a mapping 
from the set $\prgm_\al$ of all programs over $\al$ into the set $\dlp_{\al^*}$ 
of all disjunctive logic program over $\al^*=\al\cup\al_\lab\cup\bar{\al}$.
More specifically, 
\[
\tr_1:\prgm_\al\rightarrow\nnfprgm_\al
\]
translates any nested logic
program over $\al$ into a nested program in HT-NNF. 
Translation
\[\tr_2:\nnfprgm_\al\rightarrow\gdlpht_{\al\cup\al_\lab}
\]
takes these programs and transforms their rules into simpler ones as 
described by Step~2, introducing new labels. 
These rules are then fed into mapping 
\[
\tr_3:\gdlpht_{\al\cup\al_\lab}\rightarrow\gdlp_{\al\cup\al_\lab},
\] 
yielding generalised disjunctive logic programs. 
Finally, 
\[\tr_4:\gdlp_{\al\cup\al_\lab}\rightarrow\dlp_{\al^*}
\]
outputs standard disjunctive logic programs.

As argued in the following, each of these functions is PSM; 
hence, the overall function $\tr=\tr_4\circ\tr_3\circ\tr_2\circ\tr_1$ 
is PSM as well. 

We continue with the technical details, starting with $\tr_1$.

For the first step, we use the procedure $\nnf{\cdot}$ from 
Proposition~\ref{prop:NNF} to transform heads and bodies 
of rules into HT-NNF.
%
%%%%%%%%%%%%%%%%%%%%%%%%%%%%%%%%%%%%%%%%%%%definition
\begin{definition}
The function $\tr_1:\prgm_\al\rightarrow\nnfprgm_\al$ is defined by setting
\[
\tr_1(\Pi)=\{ \nnf{H(r)}\LPif\nnf{B(r)}\mid r\in\Pi \},
\]
for any $\Pi\in\prgm_\al$.
\end{definition}

Since, for each expression $\phi$, $\nnf{\phi}$ is constructible 
in polynomial time and $\phi$ is HT-equivalent to $\nnf{\phi}$ 
(cf.~Proposition~\ref{prop:NNF}), the following result is immediate:

%%%%%%%%%%%%%%%%%%%%%%%%%%%%%%%%%%%%%%%%%%%%%%%%%theorem
\begin{lemma}\label{thm:step1}
The translation $\tr_1$ is PSM.
\end{lemma}

The second step is realised as follows:

%%%%%%%%%%%%%%%%%%%%%%%%%%%%%%%%%%%%%%%%%%%%%%%%definition
\begin{definition}
The function $\tr_2:\nnfprgm_{\al}\rightarrow\gdlpht_{\al\cup\al_\lab}$ is defined by setting, for any $\Pi\in\nnfprgm_\al$, 
\[
\tr_2(\Pi)=\{\labf{H(r)}\LPif\labf{B(r)}\mid r\in\Pi\}\cup \aux{\Pi},
\]
where 
$\aux{\Pi}$ is constructed as follows:
\begin{enumerate}
\item 
for each HT-literal $l$
occurring in\/ $\Pi$, add the two rules
\[
\labf{l}\LPif l
\quad \mbox{ and } \quad 
l\LPif\labf{l};
\]

\item  
for each  expression $\phi=(\phi_1\AND\phi_2)$
occurring in\/ $\Pi$, add the three rules
\[
\labf{\phi}\LPif \labf{\phi_1}\AND\labf{\phi_2}
, \quad 
\labf{\phi_1}\LPif \labf{\phi}, 
\quad 
\labf{\phi_2}\LPif \labf{\phi} ;
\]
and

\item   
for each  expression $\phi=(\phi_1\OR\phi_2)$
occurring in\/ $\Pi$, add the three rules 
\[
\labf{\phi_1}\OR\labf{\phi_2}\LPif \labf{\phi}
, \quad
\labf{\phi}\LPif \labf{\phi_1}, \quad 
\labf{\phi}\LPif \labf{\phi_2}.
\]
\end{enumerate}
\end{definition}

This definition is basically an adaption of a structure-preserving 
normal form translation for intuitionistic logic, as described 
in \cite{Mints94}.

It is quite obvious that $\tr_2$ is modular and, for each 
$\Pi\in\nnfprgm_{\al}$, we have that $\tr_2(\Pi)$ is 
constructible in polynomial time.
In order to show that $\tr_2$ is strongly faithful, we define 
a suitable  HT-homomorphism as follows.

%%%%%%%%%%%%%%%%%%%%%%%%%%%%%%%%%%%%%%%%%%%%%%%%%%%theorem
\begin{sublemma}\label{sublemma:tr2}
Let $\tr_2$ be the translation defined above, and let 
$\tr_2^\ast:\prgm_{\al}\rightarrow\prgm_{\al\cup\al_\lab}$ 
result from $\tr_2$ by setting $\tr_2^\ast(\Pi)=\tr_2(\Pi)$ 
if\/ $\Pi\in\nnfprgm_{\al}$ and $\tr_2^\ast(\Pi)=\Pi$ if\/
$\Pi\in\prgm_{\al}\setminus\nnfprgm_{\al}$.

Then,  the function 
$\alpha_{\tr_2^\ast}:\INT{\al}\times\prgm_{\al}\rightarrow\INT{\al\cup\al_\lab}$,
defined as
\[
\alpha_{\tr_2^\ast}(\F,\Pi) = \langle I_H \cup \lambda_H(\F,\Pi), I_T \cup \lambda_T(\F,\Pi)  \rangle,
\]
is a $\tr_2^\ast$-associated HT-homomorphism,
where 
$$
\lambda_w(\F,\Pi) 
=
\{\labf{\phi}\in\al_\lab
\cap \var{\tr_2^\ast(\Pi)} 
\mid
\valF{w}{\phi}=1\}
$$
if\/ $\Pi\in\nnfprgm_{\al}$, and $\lambda_w(\F,\Pi)=\emptyset$ otherwise, 
for any $w\in\{H,T\}$ and any 
HT-interpretation $\F=\Iht$ over~$\al$.
\end{sublemma}

Hence, according to Theorem~\ref{thm:strong-faithfulness}, 
$\tr_2^\ast$ is strongly faithful.
As a consequence, $\tr_2$ is strongly faithful as well.
Thus, the following holds:

%%%%%%%%%%%%%%%%%%%%%%%%%%%%%%%%%%%%%%%%%%%%%%theorem
\begin{lemma}\label{thm:delta2}
The function $\tr_2$ is PSM.
\end{lemma}

For Step~$3$, we use a method due to 
Lifschitz~\emph{et al}.~\cite{Lifschitz99} for 
eliminating double negations in heads and bodies of rules.
The corresponding function $\tr_3$ is defined as follows:

%%%%%%%%%%%%%%%%%%%%%%%%%%%%%%%%%%%%%%%%%%%%%%definition
\begin{definition}
Let
$\tr_3:\gdlpht_{\al\cup\al_\lab}\rightarrow\gdlp_{\al\cup\al_\lab}$ 
be the function obtained by replacing, for each given 
program
$\Pi\in\gdlpht_{\al\cup\al_\lab}$, each rule $r\in\Pi$ of form
\[
\phi \OR \neg\neg p \LPif \psi \quad \mbox{\ by\ }\quad
\phi \LPif \psi \AND \neg p,
\]
as well as each rule of form
\[
\phi \LPif \psi \AND \neg\neg q \quad \mbox{\ by\ }\quad
\phi\OR \neg q \LPif \psi,
\]
where $\phi$ and $\psi$ are expressions and $p,q\in\al$.
\end{definition}

As shown in \cite{Lifschitz99}, performing replacements 
of the above type results in programs which are 
strongly equivalent to the original programs.
In fact, it is easy to see that such replacements yield 
transformed programs which are strongly faithful to the 
original ones.
Since these transformations are clearly modular and 
constructible in polynomial time, we obtain that $\tr_3$ is PSM.

%%%%%%%%%%%%%%%%%%%%%%%%%%%%%%%%%%%%%%%%%%%%%theorem
\begin{lemma}\label{thm:delta3}
The function $\tr_3$ is PSM.
\end{lemma}

Finally, we eliminate remaining negations possibly occurring in the heads of rules.
To this end, we employ a procedure due to Janhunen~\cite{Janhunen01}.

%%%%%%%%%%%%%%%%%%%%%%%%%%%%%%%%%%%%%%%%%%%%%%%%%%%%%%%%%%%%%%%%%%%%%%%%%%%%
%  \tr_4
%%%%%%%%%%%%%%%%%%%%%%%%%%%%%%%%%%%%%%%%%%%%%%%%%%%%%%%%%%%%%%%%%%%%%%%%%%%%
\begin{definition}\label{def:jan}
Let
$\tr_4:\gdlp_{\al\cup\al_\lab}\rightarrow\dlp_{\al\cup\al_\lab\cup\bar{\al}}$ 
be the function defined by setting, for any program $\Pi\in\gdlp_{\al\cup\al_\lab}$,
\begin{eqnarray*}
\tr_4(\Pi)  & \!\!\!=\!\!\! & \bar{\Pi} \cup \{
\bot\LPif(p\AND\overline{p}),\,\,\overline{p}\LPif\neg p\mid 
\neg p \mbox{ occurs in}\\
&&\hphantom{\Pi\cup\{}\mbox{the head of some rule in } \Pi\}, 
\end{eqnarray*}
where 
$\bar{\Pi}$ results from $\Pi$ by 
replacing each occurrence of a literal 
 $\neg p$ in the head of a rule in $\Pi$ by $\overline{p}$.
\end{definition}

Janhunen showed that replacements of the above kind lead to a 
transformation which is PFM.
As a matter of fact, since his notion of faithfulness is somewhat 
stricter than ours, the results in~\cite{Janhunen01} actually 
imply that, for any $\Pi,\Pi'\in\gdlp_{\al\cup\al_\lab}$, 
$\asal{\al\cup\al_\lab}{\Pi\cup\Pi'}$ is given by
\[
\{I\cap(\al\cup\al_\lab)\mid 
I \in\asal{\al\cup\al_\lab\cup\bar{\al}}{\tr_4(\Pi)\cup\Pi'}\}.
\]
However, we need a stronger condition here, viz.\ that 
the above equation holds for any $\Pi\in\gdlp_{\al\cup\al_\lab}$ 
and any $\Pi'\in\prgm_{\al\cup\al_\lab}$.
We show this by appeal to Theorem~\ref{thm:strong-faithfulness}.

%%%%%%%%%%%%%%%%%%%%%%%%%%%%%%%%%%%%%%%%%%%%%%%%%%%theorem
\begin{sublemma}
Let $\tr_4$ be the translation defined above, and let 
$\tr_4^\ast:\prgm_{\al\cup\al_\lab} 
\rightarrow\prgm_{\al\cup\al_\lab\cup\bar{\al}}$ result from 
$\tr_4$ by setting $\tr_4^\ast(\Pi)=\tr_4(\Pi)$ if\/ 
$\Pi\in\gdlp_{\al\cup\al_\lab}$ and $\tr_4^\ast(\Pi)=\Pi$ 
if\/ $\Pi\in\prgm_{\al\cup\al_\lab}\setminus\gdlp_{\al\cup\al_\lab}$.

Then,  the function 
$\alpha_{\tr_4^\ast}:\INT{\al\cup\al_\lab}\times
\prgm_{\al\cup\al_\lab}\rightarrow\INT{\al\cup\al_\lab\cup\bar{\al}}$,
defined as
\[
\alpha_{\tr_4^\ast}(\F,\Pi)  = \langle I_H \cup \kappa(\F,\Pi), 
I_T \cup  \kappa(\F,\Pi) \rangle,
\]
is a $\tr_4^\ast$-associated HT-homomorphism, 
where 
\[
\begin{array}{r@{}l}
\kappa(\F,\Pi) =  \{ \overline{p} \mid & \neg p 
\mbox{ occurs in
the head of some rule in\/ $\Pi$ }\\
&\mbox{ and } 
p\notin I_T \} 
\end{array}
\]
if\/ $\Pi\in\gdlp_{\al\cup\al_\lab}$, and 
$\kappa(\F,\Pi)=\emptyset$ 
otherwise,
for any 
 HT-interpretation $\F=\Iht$ over $\al\cup\al_\lab$.
\end{sublemma}

Observe that, in contrast to the definition of function 
$\alpha_{\tr_2^\ast}$ from Sublemma~\ref{sublemma:tr2}, 
here the same set of newly introduced atoms is
added to both worlds.
As before, we obtain that $\tr_4^\ast$ is strongly faithful, 
and hence that $\tr_4$ is strongly faithful as well.

%%%%%%%%%%%%%%%%%%%%%%%%%%%%%%%%%%%%%%%%%%%%%%%%%theorem
\begin{lemma}\label{thm:delta4}
The function $\tr_4$ is PSM.
\end{lemma}

Summarising, we obtain our main result, which is as follows:

%%%%%%%%%%%%%%%%%%%%%%%%%%%%%%%%%%%%%%%%%%%%%%%%%theorem
\begin{theorem}
Let $\tr_1\commadots\tr_4$ be the functions defined above.
Then, the composed function 
$\tr=\tr_4\circ\tr_3\circ\tr_2\circ\tr_1$, mapping nested 
logic programs over alphabet $\al$ into disjunctive logic 
programs over alphabet $\al\cup\al_\lab\cup\bar{\al}$, 
is polynomial, strongly faithful, and modular.
\end{theorem}

Since strong faithfulness implies faithfulness, we get the following corollary:

%%%%%%%%%%%%%%%%%%%%%%%%%%%%%%%%%%%%%%%%%%%%%%%%%%%%theorem
\begin{corollary}
For any nested logic program $\Pi$ over $\al$, the answer 
sets of\/ 
$\Pi$ are in a one-to-one correspondence to the answer 
sets of $\tr(\Pi)$, determined by the following equation:
\[
\asal{\al}{\Pi}  =  \{ I \cap \al \mid I \in \asal{\al^\ast}{\tr(\Pi)}\},
\]
where $\al^\ast=\al\cup\al_\lab\cup\bar{\al}$.
\end{corollary}

We conclude with a remark concerning the construction 
of function $\tr_2$.
As pointed out previously, this mapping is based on a 
structure-preserving normal form translation for 
intuitionistic logic, as described in \cite{Mints94}.
Besides the particular type of translation used here, 
there are also other, slightly improved structure-preserving 
normal form translations in which fewer rules are introduced, 
depending on the polarity of the corresponding subformula 
occurrences.
However, although such optimised methods work in monotonic 
logics, they are not sufficient in the present setting.
For instance, in a possible variant of translation $\tr_2$
based on the polarity of subformula occurrences,
instead of introducing all three rules for 
an expression $\phi$ of form $(\phi_1\AND\phi_2)$,
 only 
$\labf{\phi}\LPif \labf{\phi_1}\AND\labf{\phi_2}$
is used if $\phi$ occurs in the body of some rule, or 
both $\labf{\phi_1}\LPif \labf{\phi}$
and 
$\labf{\phi_2}\LPif \labf{\phi}$
are used if $\phi$ occurs in the head of some rule, and 
analogous manipulations are performed
for atoms and disjunctions.
Applying such an encoding to 
$\Pi=\{ p \LPif; \; q \LPif; \; r\OR(p\AND q)\LPif \ \}$ 
over $\al_0=\{p,q,r\}$ yields a translated program possessing 
two answer sets, say $S_1$ and $S_2$, such that 
$S_1\cap\al_0 = \{p, q\}$ and $S_2\cap\al_0 =\{p,q,r\}$, 
although only $\{p,q\}$ is an answer set of $\Pi$.

\section{Conclusion}

We have developed a translation of logic programs with nested expressions into
disjunctive logic programs.
We have proven that our translation is 
polynomial,
strongly faithful,
and modular.
This allows us to utilise off-the-shelf disjunctive logic programming systems
for interpreting nested logic programs.
In fact, we have implemented our translation as a front end 
for the system \DLV~\cite{Eiter97,Eiter98a}.
The corresponding compiler is implemented in Prolog and can be downloaded from the
Web at URL
\begin{center}
\begin{small}
\texttt{http://www.cs.uni-potsdam.de/$\sim$torsten/nlp}.
\end{small}
\end{center}

Our technique is based on the introduction of new atoms, 
abbreviating subformula occurrences.
This method has its roots in structure-preserving normal form
translations~\cite{Tseitin70,Greenbaum86}, which are frequently  used in
automated deduction.
In contrast to theorem proving applications, however, where the main focus is to
provide satisfiability (or, alternatively, validity) preserving translations,
we are concerned with much stronger equivalence properties, 
involving additional minimality criteria,
since our goal
is to \emph{reconstruct} the answer sets of the original programs from the
translated ones.

With the particular labeling technique employed here, our translation 
avoids the risk of an exponential
blow-up in the worst-case,
faced by a previous approach of 
Lifschitz \emph{et al.}~\cite{Lifschitz99} due to
the usage of distributivity laws.
However, this is not to say that our translation is \emph{always} the better choice.
As in classical theorem proving, it is rather a matter of experimental studies under
which circumstances which approach is the more appropriate one.
To this end, besides the implementation of our structural translation,
we have also implemented the distributive translation into
disjunctive logic programs in order to conduct experimental results.
These experiments are subject to current research.

Also, we have introduced the concept of \emph{strong faithfulness},
as a generalisation of (standard) faithfulness and strong equivalence.
This allows us, for instance, to translate, in a semantics-preserving way, 
arbitrary program parts and leave the remaining program unaffected.

\end{document}